\documentclass[a4paper,conference]{IEEEtran}
\usepackage{graphicx}
\usepackage{amsmath}
\usepackage{amssymb}
\usepackage{booktabs}

\usepackage[utf8]{inputenc} 
\usepackage{url}            
\usepackage{booktabs}       
\usepackage{amsfonts}       
\usepackage{nicefrac}       
\usepackage{microtype}      
\usepackage{lipsum}
\usepackage{xcolor}
\usepackage{multirow}
\usepackage{array}
\usepackage{pifont}
\usepackage{rotating} 
\usepackage{amssymb}
\usepackage{amsmath}
\usepackage{tabularx}
\usepackage{algorithm}
\usepackage{algorithmic}
\usepackage[switch]{lineno}
\graphicspath{ {./images/} }
\usepackage{listings}

\usepackage{amsmath,amsfonts,bm}

\def\eqref#1{equation~\ref{#1}}

\def\1{\bm{1}}

\def\vc{{\bm{c}}}

\DeclareMathAlphabet{\mathsfit}{\encodingdefault}{\sfdefault}{m}{sl}
\SetMathAlphabet{\mathsfit}{bold}{\encodingdefault}{\sfdefault}{bx}{n}

\newcommand{\Cmat}{{\bf C}}

\newcommand{\pv}{{\boldsymbol p}}

\newcommand{\yv}{{\boldsymbol y}}

\newcommand{\Lcal}{\mathcal{L}}

\newcommand{\Rcal}{\mathcal{R}}

\newcommand{\Ep}{{\mathbb E}}

\ifCLASSINFOpdf
\else
\fi
\begin{document}
\IEEEoverridecommandlockouts
%
\title{Augmenting Vision Language Pretraining by Learning Codebook with Visual Semantics}



%
\author{\IEEEauthorblockN{$^*$Xiaoyuan Guo\IEEEauthorrefmark{2},
$^*$Jiali Duan\thanks{$^*$ Guo and Duan contributed equally to the work.}\IEEEauthorrefmark{3},
C.-C. Jay Kuo\IEEEauthorrefmark{3},
Judy Wawira Gichoya\IEEEauthorrefmark{4} and
Imon Banerjee\IEEEauthorrefmark{5},\IEEEauthorrefmark{6}}
\IEEEauthorblockA{\IEEEauthorrefmark{2}Department of Computer Science, Emory University, GA, USA}
\IEEEauthorblockA{\IEEEauthorrefmark{3}Ming Hsieh Department of Electrical and Computer Engineering, University of Southern California, CA, USA}
\IEEEauthorblockA{\IEEEauthorrefmark{4}Department of Radiology and Imaging Sciences, Emory University, GA, USA}
\IEEEauthorblockA{\IEEEauthorrefmark{5}School of Computing, Informatics, and Decision Systems Engineering, Arizona State University, AZ, USA}
\IEEEauthorblockA{\IEEEauthorrefmark{6}Department of Radiology, Mayo Clinic, AZ, USA
\\ Email: xiaoyuan.guo@emory.edu, jialidua@usc.edu, cckuo@sipi.usc.edu, judywawira@emory.edu, banerjee.imon@mayo.edu}}


\maketitle

\begin{abstract}
Language modality within the vision language pretraining framework is innately
discretized, endowing each word in the language vocabulary a semantic meaning. In contrast, visual modality is inherently continuous and high-dimensional, which potentially prohibits the alignment as well as fusion between vision and language modalities. We therefore propose to ``discretize'' the visual representation by joint learning a codebook that imbues each visual token a semantic. We then utilize these discretized visual semantics as self-supervised ground-truths for building our Masked Image Modeling objective, a counterpart of Masked Language Modeling which proves successful for language models. To optimize the codebook, we extend the formulation of VQ-VAE which gives a theoretic guarantee. Experiments validate the effectiveness of our approach across common vision-language benchmarks. 
\end{abstract}


%
\IEEEpeerreviewmaketitle

\section{Introduction}~\label{intro}
Inspired by the success of language modeling~\cite{brown2020language,devlin2018bert}, the concept of Vision-Language Pretraining (V\&L) has attracted growing attention in the community, where the model is pretrained once and achieves superior performance over a set of downstream tasks, via transfer learning. The central theme to the problem is learning better alignment and interactions between the two modalities via feature fusion. Two popular ways - late fusion and early fusion have been researched. Late fusion approaches such as CLIP~\cite{radford2021learning} and ALIGN~\cite{jia2021scaling} directly optimize the InfoNCE~\cite{oord2018representation} objective, by leveraging large amount of paired data (400M for CLIP and 1.8B for ALIGN).  On the other hand, early fusion approaches such as  VinVL~\cite{zhang2021vinvl}, ViLT~\cite{kim2021vilt} and OSCAR~\cite{li2020oscar} adopt a multi-modal transformer to model the interactions between the vision and text modalities. On top of which, objectives such as image-text matching (ITM) and masked language modeling (MLM) are optimized to enforce the alignment. Our work belongs to this category.

MLM has proven successful for language modeling. The key idea is to predict masked text tokens given an image and unmasked text tokens. Questions naturally arise, \textit{can we do the same for the image modality? Will it be as effective as MLM for performance?} The main hurdle for applying masked image modeling (MIM) lies in the difference between vision and language. Language is inherently discrete, endowing each word in the language vocabulary a semantic meaning. In contrast, visual modality is continuous and high-dimensional. Existing approaches such as VinVL~\cite{zhang2021vinvl}, OSCAR~\cite{li2020oscar}, UNITER~\cite{chen2020uniter} utilize an object detector to assign a class ID to the visual patches, based on which the MIM objective is built. However, the number of classes that an object detector can recognize is limited and the detector is not end-to-end optimized.

To overcome the challenge, we propose a \textbf{C}ode\textbf{B}ook based approach for \textbf{Vi}sion \textbf{L}anguage \textbf{A}lignment (\textbf{CB-ViLA}), which can help quantize visual features and facilitate optimizing image-text alignment as well as the MIM objective. Inspired by the ability of learning discrete visual representations of VQ-VAE~\cite{oord2017neural}, we extend its formulation into the multi-modal setting to learn a codebook with visual semantics. 
Specifically, we approximate the latent space posterior $q(z|x)$ with a visual encoder and the codebook is then used to ``discretize" the latent space into a probability distribution over the codebook vectors $q(z=c_k|x)$. During decoding, we incorporate the language modality into the VAE reconstruction objective by taking text sequence features as a conditional variable. In this way, our multi-modal fusion encoder can also be interpreted as VQ-VAE decoder, where its parameters are shared for multi-modal interactions and VQ-VAE decoding (See Figure~\ref{fig:framework}). 

Although codebook is also utilized in works such as BEiT~\cite{bao2021beit}, DALLE~\cite{ramesh2021zero}, they are different from our design. As the codebook used in BEiT~\cite{bao2021beit} is off-the-shelf from DALLE~\cite{ramesh2021zero}, kept frozen during pretraining. While in SOHO~\cite{huang2021seeing}, the codebook is heuristically updated via momentum. Instead, we integrate codebook learning into the vision-language pretraining framework by alternately optimizing the codebook and the encoders (Bert~\cite{devlin2018bert}, ViT~\cite{dosovitskiy2020image} and multi-modal fusion encoder). The codebook gradient is frozen when computing MIM while the encoders are updated.

To summarize, our main contributions are as follows:
\begin{enumerate}
    \item We present a vision language framework that unifies MLM and MIM by jointly optimizing a visual semantic codebook, on the evidence lower
    bound of language-conditioned pixel reconstruction posterior.
    \item  We show that the quantization codebook is able to learn useful visual semantics, and together with the MIM objective, help improve the performance of the framework over downstream tasks. 
\end{enumerate}

\section{Related Work}
\textbf{Vision-Language Pretraining.} V\&L has been popular recently as it enables transfer of superior performance in a wide variety of downstream vision-language tasks by pre-training once, similar to language modelling~\cite{brown2020language,devlin2018bert,chen2020uniter}. Existing works can be categorized into one-stream v.s. two-stream.
In one stream model, features of different modalities are directly fed into a transformer~\cite{su2019vl}. Whereas in two stream models, inputs are first processed by two single-modal networks before fed into a transformer~\cite{lu2019vilbert}. More recently, some works such as ALBEF~\cite{li2021align}, CODIS~\cite{duan2022multi}, TCL~\cite{yang2022vision} adopt a hybrid architecture, which combines both one-stream muti-modal encoder with two-stream uni-modal encoders. Our architecture falls into this category. The main difference between ALBEF~\cite{li2021align} and ours is visual semantic codebook learning and masked image modeling objective designed to address the discrepancy between image and text modalities.

\textbf{Codebook Usage.} Some recent works have adopted codebook for vision language pretraining. For example, CODIS~\cite{duan2022multi} proposes a multi-modal codebook as a bridge to align image and text modalities at a cluster level. The multi-modal codebook is optimized with optimal transport. The codebook in SOHO~\cite{huang2021seeing} discretizes features from the convolutional encoder and is momentum updated. Although we also focus on learning a visual quantization codebook for multi-modal alignment, the learning process is different as our codebook is updated by optimizing the evidence lower bound of language-conditioned pixel reconstruction posterior. In this sense, our framework can also be seen as an extension of VQ-VAE~\cite{oord2017neural} where ViT and multi-modal fusion are treated as encoder and decoder respectively. The idea of our codebook shares some similarity with the concept ``vector quantization'' of VideoBert~\cite{sun2019videobert}, which is achieved via hierarchical k-means with human inspection. Other than the applications in V\&L, codebook has be utilized for vision tasks~\cite{duan2021slade,bao2021beit} as well. For example, BEiT~\cite{bao2021beit}, VIOLET~\cite{fu2021violet} both use a pretrained codebook for input space tokenization. These visual codebooks are kept fixed during training. To further improve the quality of visual codebook, PeCo~\cite{dong2021peco} extends DALLE~\cite{ramesh2021zero} with a perceptual loss. Notably, our codebook is joint optimized and updated to tokenize the feature space. 



\section{Method}
\begin{figure}[!t]
  \centering
   \includegraphics[width=\linewidth]{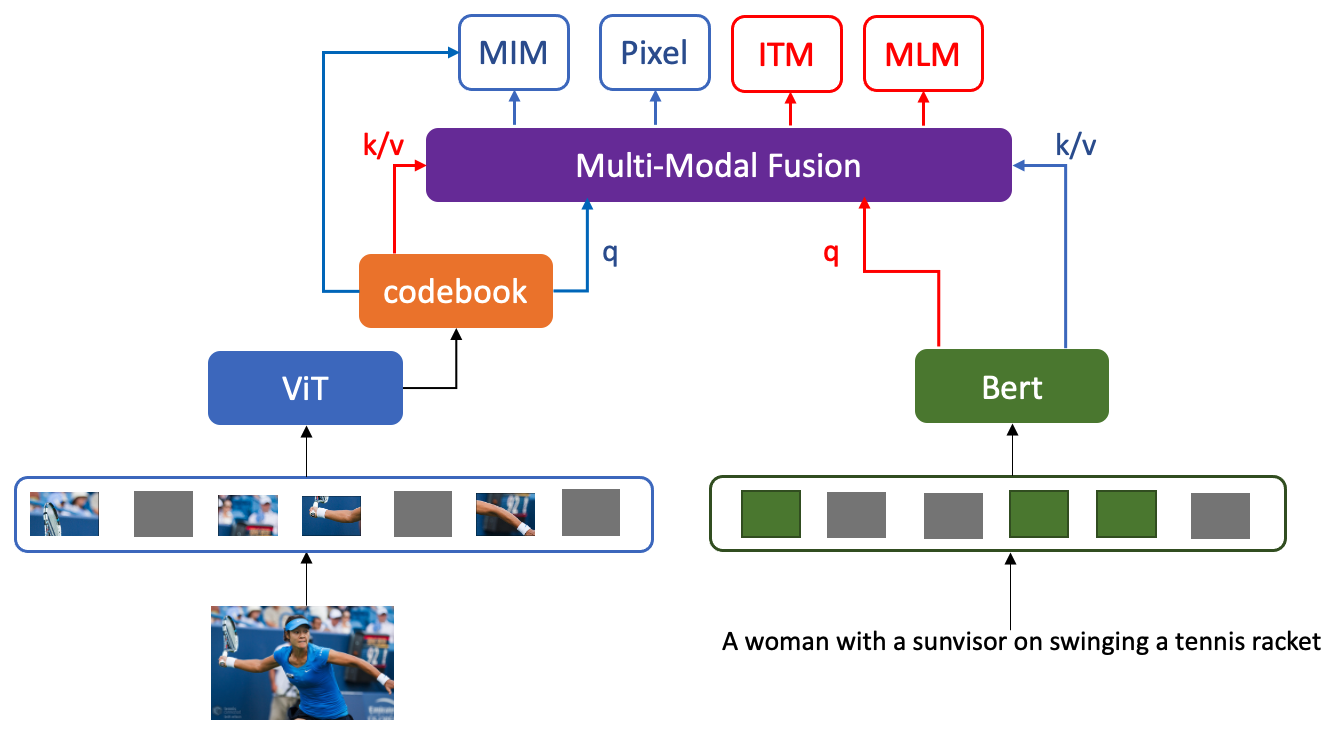}
   \caption{Overview of \textbf{CB-ViLA} framework. We leverage a codebook to enforce visual semantic alignment, via the use of text-conditioned masked image modeling. Arrows of the same color indicate the flows for a specific pretraining objective. For example, MIM is calculated with codeword indices and the \textit{cls} token of multi-modal fusion module, which is a cross-attention of ``discretize'' ViT~\cite{dosovitskiy2020image} output as query and Bert~\cite{devlin2018bert} as key/value. Our codebook is optimized via an extension of VQ-VAE~\cite{oord2017neural} objective.}  
   \label{fig:framework}
   \vspace{-2mm}
\end{figure}

Figure~\ref{fig:framework} is an overview of our \textbf{CB-ViLA} framework. Our goal is to learn visual semantics via a codebook, facilitating the alignment between image and text modalities. We describe visual semantic codebook learning in Section~\ref{sec:codebook}. In Section~\ref{sec:mim}, we explain how the codebook is integrated. In particular, the ``discretized'' visual information will be used as MIM targets and multi-modal queries respectively in a self-supervised manner. Finally, we illustrate how our proposed components fit into the vision-language pretraining framework as well as training procedure.

\subsection{Visual Semantic Codebook Learning}
\label{sec:codebook}
As mentioned in Section~\ref{intro}, language is inherently discrete and semantic rich, whereas the visual signal is continuous and could be noisy. To facilitate the alignment between the two, we leverage a learnable codebook to assign a visual semantic to each visual token. 

We denote the learnable codebook as $\Cmat =\{\vc_1, \vc_2,\ldots, \vc_K\}\in \Rcal^{d_c \times K}$, where $d_c$ is the dimension for each code and $K$ equals to the number of codewords (e.g., $3$K). Each $\vc \in \Cmat$ corresponds to a visual semantic (Refer to Section~\ref{sec:codbook_vis} for visualization). Given an encoded visual sequence $\{v^1, v^2, ..., v^N\}$, where $N$ denotes the length of visual sequence, the codebook ``discretizes'' them by doing a nearest neighbor look-up using the shared embedding space $\Rcal^{d_c \times K}$, as shown in Equation~\ref{eqn:lookup},

\begin{equation}
\label{eqn:lookup}
    z_q(v^i) = \vc_k, \; k = q(v^i) = \textrm{argmin}_k||v_i - \vc_k||_2
\end{equation}
where $k$ ($k\in[1,K]$) indexes the closest cluster (w.r.t., $l_2$ distance) that $v_i$ ($i\in[1, N]$) belongs to. The corresponding output after visual discretization is $\{z_q(v^1), z_q(v^2),...,z_q(v^N)\}$. 

To learn the codebook within the framework, we consider our latent codebook space $z$ as a random variable, and we assume a standard gaussian prior $p(z)$ over the latents. Given an image $x$, we also define a posterior for the latent, expressed in the bayes form, 
\begin{equation}
\label{eqn:bayes}
    p(z|x) = \frac{\int_{l}p(x|z,l)p(z)}{p(x)}
\end{equation}
where $l$ represents the language embedding space. As $p(x)$ is computationally intractable, we instead optimize a restricted family of independent gaussians $q(z|x)$. In our case, $q(z|x)$ is approximated by the visual encoder and $p(x|z,l)$ is represented by a multi-modal fusion encoder in Figure~\ref{fig:framework}. By maximizing the log-likelihood of $p(x|z,l)$, we get a VAE objective. However, as the look-up operation is ``discrete'' and  non-differentiable, we extend a discretized VQ-VAE objective~\cite{oord2017neural} and adopt gumbel-softmax~\cite{jang2016gumbel} for gradient back-propagation.

\begin{equation}
\label{eqn:codebook}
\begin{split}
    \Lcal_{codebook} = &\Ep( -\log(p(x|z_q(x), l) 
    + ||\textrm{sg}[\textrm{ViT}(x)] - \vc||^2_{2} \\ &+ \beta||\textrm{ViT}(x) - \textrm{sg}[\vc]||^2_2 )
\end{split}
\end{equation}

\noindent \noindent where $\vc$ can be considered as tokenization of visual input. Each element $\vc_i$ is patchwise codeword representation for patch $v^i$. $\textrm{sg}$ stands for stop-gradient. The first term corresponds to the pixel loss in Figure~\ref{fig:framework}. The second and third term are essentially enforcing a bi-directional mapping: learning codebook vectors that align to the encoder outputs and learning encoder outputs that align to codebook vector. 

\subsection{Masked Image Modeling via Codebook}
\label{sec:mim}
With the codebook introduced in the previous section, we build a counterpart of masked language modeling (MLM) in the BERT objective on top of ``discrete'' visual tokens, which we call masked image modeling (MIM). As shown in Figure~\ref{fig:framework}, an input image $x$ is first patchified into a sequence of visual patches $\{x^1,x^2,...,x^N\}$ before fed into the visual encoder $ViT$. These encoded visual tokens are classified into masked $V$ and unmasked tokens $\hat{V}$, used to calculate the MIM loss. Denote encoded text tokens as $T = \{t^1, t^2, ..., t^M\}$ (M is the length of the text), 
\begin{equation}
\label{eqn:mim}
    \Lcal_{\text{mim}} = -\Ep \log p(q(V) \, | \, l(T), z_q(\hat{V}))
\end{equation}
where the masked token index $q(V)$ is predicted based on unmasked visual tokens and the text embeddings. The supervision signal for training $\Lcal_{\text{mim}}$ comes from the codebook indices for the masked visual tokens. 

\subsection{Vision Language Pretraining}
\label{sec:pretrain}
Two commonly used objectives for multimodal training frameworks are: (i) masked language modeling loss (MLM) and (ii) image-text matching (ITM), which we build on top of multi-modal fusion encoder. 

\noindent\textbf{Image-Text Matching (ITM) Loss } 
Given an arbitrary pair of image and text, ITM predicts whether they are aligned (positive pairs) or not (negative pairs). This procedure can be formulated as a binary classification problem.
Specifically, [CLS] token from the fusion encoder is used as the joint representation of the image-text pair. ITM head is a fully connected layer to predict the matching probability $\pv_{\text{itm}}$. 
We assume that each image-text pair $(I_i, T_i)$ sampled from the pre-training datasets is a positive example and construct negative examples through the following strategy:
for each image $I_i$ within the batch, we sample one negative text $T_j$ from the same batch based on the contrastive similarity distribution. So that text that is more similar to this image will have a higher chance to get sampled. Similarly, one hard negative image will be sampled for each text $T_i$.
We denote $y_{\text{itm}}$ as the ground-truth labels indicating whether the image-text pair is positive or negative.
\begin{equation}
    \Lcal_{\text{itm}} = -\Ep_{I,T\sim \pv_\text{data}} H(\pv_{\text{itm}}, \yv_{\text{itm}})
\end{equation}
where $H$ is the cross entropy operator.

\noindent\textbf{Masked Language Modeling (MLM) Loss } We follow the design of MLM loss from Bert \cite{devlin2018bert}, which aims to predict the ground-truth labels of masked text tokens $y_{\text{mlm}}$.
Specifically, we randomly mask out 15\% of input text tokens, those masked tokens are replaced with special token [MASK].
Different from Bert, our MLM loss is conditioned on both surrounding text tokens and image representations.
Assume the predicted token probability is $\pv_{\text{mlm}}$, we construct the loss objective as follows, 

\begin{equation}
    \Lcal_{\text{mlm}} = -\Ep_{I,\hat{T}\sim \pv_\text{data}} H(p_{\text{mlm}}, \yv_{\text{mlm}})
\end{equation}
where $\hat{T}$ is the text token sequence after masking.

In summary, we simultaneously optimize the codebook and the dual unimodal encoders within the framework in an end-to-end manner, employing the losses discussed in previous sections as follows, 
\begin{equation}
    \Lcal_{\text{final}} = \Lcal_{\text{codebook}} + \Lcal_{\text{mim}} + \Lcal_{\text{itm}}  + \Lcal_{\text{mlm}}
\end{equation}
among which ITM and MLM losses are calculated with text representation as queries and visual representation as key and values. The cross-attention relation is reversed for MIM and pixel losses. The pixel loss is a part of codebook loss outlined in Equation~\ref{eqn:codebook}. Arrows of the same color in Figure~\ref{fig:framework} indicate the flows for calculating a corresponding objective.

\section{Experiments}
To evaluate our approach, we conduct extensive studies on commonly used benchmarks and present experimental comparisons against state-of-the-art V\&L methods as shown in this section.

\begin{table*}[!ht]
\caption{Performance comparison of \textbf{zero-shot} image-text retrieval on Flickr30K and COCO datasets.\vspace{-3mm}}
	\label{table:zero_shot}
	\footnotesize
	\setlength\tabcolsep{8pt}
	\begin{center}
		\begin{tabular}{lcccccc|cccccc}
            \toprule 
                \multirow{3}{*}{Method} & \multicolumn{6}{c}{Flickr30K (1K)} & \multicolumn{6}{c}{MSCOCO (5K)}\\
                
                & \multicolumn{3}{c}{Text Retrieval} & \multicolumn{3}{c}{Image Retrieval} &
                \multicolumn{3}{c}{Text Retrieval} & \multicolumn{3}{c}{Image Retrieval} \\
            
                & R@1 & R@5 & R@10 & R@1 & R@5 & R@10
                & R@1 & R@5 & R@10 & R@1 & R@5 & R@10 \\
                \midrule
                ImageBERT~\cite{qi2020imagebert} & 70.7 & 90.2 & 94.0 & 54.3 & 79.6 & 87.5 & 44.0 & 71.2 & 80.4 & 32.3 & 59.0 & 70.2 \\
                Unicoder-VL~\cite{li2020unicoder}  & 64.3 & 85.8 & 92.3 & 48.4 & 76.0 & 85.2 &  - & - & - & - & - & - \\
                UNITER~\cite{chen2020uniter} &  80.7 & 95.7 & 98.0 & 66.2 & 88.4 & 92.9 &  - & - & - & - & - & -  \\
                ViLT~\cite{kim2021vilt} & 73.2 & 93.6 & 96.5 &  55.0 & 82.5 & 89.8 & 56.5 & 82.6 & 89.6 &  40.4 & 70.0 & 81.1 \\
                CLIP \cite{radford2021learning} &  88.0 & 98.7 & 99.4 & 68.7 & 90.6 & 95.2 &  58.4 & 81.5 & 88.1 & 37.8 & 62.4 & 72.2 \\
                
                ALIGN \cite{jia2021scaling} &  88.6 & 98.7 &  \textbf{99.7} & 75.7 & 93.8 &  \textbf{96.8} &  58.6 & 83.0 & 89.7 & 45.6 & 69.8 & 78.6 \\
                
                ALBEF 4M \cite{li2021align} & 90.5 & 98.8 &  \textbf{99.7} & 76.8 & 93.7 & 96.7 & 68.6 & 89.5 & 94.7 & 50.1 & 76.4 & 84.5 \\
                \textbf{CB-ViLA}(Ours)  &  \textbf{91.9} &  \textbf{99.1} & 99.6 &  \textbf{79.1}&  \textbf{94.5} & 96.6 &  \textbf{70.1} &  \textbf{90.2} &  \textbf{95.3} &  \textbf{52.4} &  \textbf{77.8} &  \textbf{86.0}\\
                
                \bottomrule
            \end{tabular}
	\end{center}
	
\end{table*}
\newcommand{\xmark}{\ding{55}}
\begin{table*}[!ht]
\caption{Performance comparison of \textbf{fine-tuned} image-text retrieval on Flickr30K and COCO datasets.\vspace{-3mm}}
	\label{table:fine_tune}
	\footnotesize
	\setlength\tabcolsep{8pt}
	\begin{center}
		\begin{tabular}{lcccccc|cccccc}
            \toprule 
                \multirow{3}{*}{Method} & \multicolumn{6}{c}{Flickr30K (1K)} & \multicolumn{6}{c}{MSCOCO (5K)}\\
                
                & \multicolumn{3}{c}{Text Retrieval} & \multicolumn{3}{c}{Image Retrieval} &
                \multicolumn{3}{c}{Text Retrieval} & \multicolumn{3}{c}{Image Retrieval} \\
            
                & R@1 & R@5 & R@10 & R@1 & R@5 & R@10
                & R@1 & R@5 & R@10 & R@1 & R@5 & R@10 \\
                \midrule
                ImageBERT~\cite{qi2020imagebert} & 87.0 & 97.6 & 99.2 & 73.1 & 92.6 & 96.0 & 66.4 & 89.8 & 94.4 & 50.5 & 78.7 & 87.1 \\    
            
                UNITER~\cite{chen2020uniter} & 87.3 & 98.0 & 99.2 & 75.6 & 94.1 & 96.8 & 65.7 & 88.6 & 93.8 & 52.9 & 79.9 & 88.0 \\
                
                VILLA~\cite{gan2020large} & 87.9 & 97.5 & 98.8 & 76.3 & 94.2 & 96.8 & - & - & - & - & - & - \\
                
                OSCAR~\cite{li2020oscar} & - & - & - & - & - & - & 70.0 & 91.1 & 95.5 &  54.0 & 80.8 & 88.5 \\
                
                ViLT~\cite{kim2021vilt} & 83.5 & 96.7 & 98.6 &  64.4 & 88.7 & 93.8 & 61.5 & 86.3 & 92.7  & 42.7 & 72.9 & 83.1 \\
                
                UNIMO~\cite{li2020unimo} &  89.7 & 98.4 & 99.1 & 74.6 & 93.4 & 96.0 & - & - & - & - & - & - \\
                
                
                
                SOHO \cite{huang2021seeing} &  86.5 & 98.1 & 99.3 &  72.5 & 92.7 & 96.1 & 66.4 & 88.2 & 93.8 & 50.6 & 78.0 & 86.7 \\
                
                ALBEF 4M \cite{li2021align} & 94.3 & \textbf{99.4} & 99.8 & 82.8 & \textbf{96.7} & \textbf{98.4} &  73.1 & 91.4 & 96.0 & 56.8 & 81.5 & 89.2 \\
                
                \textbf{CB-ViLA}(Ours)  & \textbf{95.1} & 99.1 & \textbf{99.9} & \textbf{83.0} & 96.1 & 98.2 &  \textbf{76.2} & \textbf{92.3} & \textbf{97.1} & \textbf{58.1} & \textbf{83.1} & \textbf{89.6} \\
            
                \bottomrule
            \end{tabular}
	\end{center}
	
	\vspace{-0.2in}
\end{table*}

\subsection{Pre-training Datasets}
We follow previous experimental protocols~\cite{chen2020uniter,li2021align}  for fair comparisons. The pretraining datasets include COCO~\cite{lin2014microsoft}, Visual Genome (VG)~\cite{krishna2017visual}, Conceptual Captions (CC)~\cite{sharma2018conceptual}, and SBU Captions~\cite{ordonez2011im2text}. In total, there are 4.0M unique images and 5.1M image-text pairs. 

\subsection{Downstream Tasks}
\noindent\textbf{Image-Text Retrieval} This consists of two tasks: (1) image as query and text as targets (TR); (2) text as query and image as targets (IR). The pre-trained model is evaluated on COCO~\cite{lin2014microsoft} and Flickr30K \cite{plummer2015flickr30k} by following both fine-tuning and zero-shot settings. For the fine-tuning setting, the pre-trained model is fine-tuned on the training data and evaluated on the validation/test data. For the zero-shot setting, the pre-trained model is directly evaluated on the test data without any further training.

\noindent\textbf{Visual Question Answering (VQA)~\cite{goyal2017making}} This task
aims to predict the answer given an image and a question (in text format), which requires an understanding of vision, language and commonsense knowledge to answer.
We consider this task as a generation problem by following the same setting in \cite{li2021align}.

\noindent\textbf{Visual Entailment (SNLI-VE)~\cite{xie2019visual}} It
predicts whether a given image semantically entails a given text, which is a three-classes classification problem.
Specifically, the class or relationship between any given image-text pair can be entailment, neutral, or contradictory.

\noindent\textbf{Visual Reasoning (NLVR$^2$)~\cite{suhr2018corpus}} The task
determines whether a natural language caption is true about a pair of photographs.
We evaluate our model on NLVR$^2$ dataset which contains 107,292 examples of human-written English sentences paired with web photographs. 

\subsection{Implementation Details}
We use standard ViT-B/16~\cite{dosovitskiy2020image} as our vision encoder and 6-layer Bert\textsubscript{base}~\cite{devlin2018bert} as text encoder. Our multi-modal fusion encoder is another 6-layer Bert but has the same architecture as the text encoder. The multi-modal fusion encoder is shared among ITM, MLM, Pixel and MIM for forward passes and we adopt cross-attention mechanism to reduce computation. We set codebook size to 3K, as we didn't observe further improvement with larger codebook size. We follow UNITER~\cite{chen2020uniter} to set 15\% masking ratio for masked language modeling and we follow MAE~\cite{he2021masked} to set 75\% masking ratio for masked image modeling. In the pre-training stage, the model is trained for 30 epochs with a batch size of 512. We use mini-batch AdamW optimizer~\cite{loshchilov2017decoupled} with a weight decay of 0.02.
We burn in MIM and MLM objectives after warming up the codebook for 1,000 iterations by training ITM and Pixel loss objectives.
The learning rate is initialized as $1e-5$ and first warmed-up to $1e-4$ after 1,000 iterations.
Then it's decreased with a cosine decay strategy to $1e-5$. All of our experiments were performed on $8$ NVIDIA A100 GPUs, with an approximate training time of 60 hours.

\subsection{Evaluation on Image-Text Retrieval}
For the image-text retrieval tasks, we conduct two different scenarios for evaluation: ``zero-shot" retrieval task and ``after-finetuning" retrieval task, following the setting in~\cite{li2020oscar,chen2020uniter,li2021align}. We compare with both early-fusion methods such as ViLT, OSCAR, UNITER and late-fusion methods such as ~\cite{jia2021scaling,radford2018improving}. ALBEF is a hybrid approach that performs feature alignment along with fusion. We evaluate our method on Flickr30K (1K test set) and MSCOCO (5K) in Table \ref{table:zero_shot} and \ref{table:fine_tune}. In the zero-shot setting, we achieve $11.2\%/12.9\%$ TR/IR improvement compared with the early-fusion approach UNITER on Flickr30K in R@1. Compared to SOTA late-fusion approach ALIGN, our approach increases $3.3\%/3.4\%$ respectively on Flickr30K R@1 and a significant $11.5\%/6.8\%$ gain on MSCOCO R@1. We hypothesize that MSCOCO is more challenging than Flickr and thus more representative of the true performance gap. Comparing to a recent SOTA approach ALBEF (4M), we show a margin of $1.4\%/2.3\%$ in terms of R@1 for TR/IR on Flickr30K and $1.5\%/2.3\%$ R@1 for TR/IR on MSCOCO respectively. In the finetuning comparison, performance tend to converge, but we observe a similar trend in performance boost across Flickr30K and MSCOCO, especially with R@1 metrics.

\begin{table}[h]
    \vspace{1.5ex}
	\caption
	{
	\small	
		Comparison with a variety of state-of-the-art methods on downstream vision-language tasks: VQA, NVLR$^2$, SNLI-VE.\vspace{-2mm}
	}
	\label{tbl:vqa_nlvr_ve}
    \aboverulesep = 0.48mm
    \belowrulesep = 0.48mm
    \small
    \footnotesize
	\centering	
	\resizebox{0.5\textwidth}{!}{%
	\begin{tabular}	{l   |  c  c  c  c  c  c  }
		\toprule	 	
	 \multirow{2}{*}{Method} & \multicolumn{2}{c}{VQA} & \multicolumn{2}{c}{NLVR$^2$} & \multicolumn{2}{c}{SNLI-VE} \\
	  & {test-dev} & {test-std} & {dev} & {test-P} & {val} & {test}\\
	  \midrule
	  VisualBERT~\cite{li2019visualbert} & 70.80 & 71.00 & 67.40 & 67.00 & - & - \\
	  VL-BERT~\cite{lu2019vilbert} & 71.16 & - &  - & - & - & - \\
	  LXMERT~\cite{tan2019lxmert}  & 72.42 & 72.54 & 74.90 & 74.50 & - & - \\
	  12-in-1~\cite{lu202012} & 73.15 & - & - & 78.87 & - & 76.95 \\
	  UNITER~\cite{chen2020uniter} & 72.70 & 72.91 & 77.18 & 77.85 & 78.59 & 78.28 \\
	   VL-BART/T5~\cite{cho2021unifying} & - & 71.3 & - & 73.6& - & - \\
	   ViLT~\cite{kim2021vilt}  & 70.94 & - & 75.24 & 76.21& - & - \\
	   OSCAR~\cite{li2020oscar} &  73.16 & 73.44 & 78.07 & 78.36 & - & - \\
	   VILLA~\cite{gan2020large} & 73.59 & 73.67 & 78.39 & 79.30 & 79.47 & 79.03 \\
	  ALBEF 4M\cite{li2021align}  & 74.54 & \textbf{74.70} & 80.24 & 80.50 & 80.14 & 80.30 \\
	  \midrule
	   \textbf{CB-ViLA}(Ours) & \textbf{75.84} & 74.20 & \textbf{80.50} &  \textbf{80.84}& \textbf{81.47} &  \textbf{81.40} \\
		\bottomrule
	\end{tabular}
 	}\vspace{-3mm}
\end{table}		
\begin{table}[ht]
	\caption{Performance comparison of zero-shot image-text retrieval on COCO datasets for ablation studies.\label{table:ablation}\vspace{-3mm}}
	\footnotesize
	\setlength\tabcolsep{8pt}
	\begin{center}
	\resizebox{\linewidth}{!}{
		\begin{tabular}{ccccccc}
            \toprule 
                \multirow{3}{*}{Objective functions} & \multicolumn{6}{c}{MSCOCO (5K)} \\
                & \multicolumn{3}{c}{Text Retrieval} & \multicolumn{3}{c}{Image Retrieval} \\
                & R@1 & R@5 & R@10 & R@1 & R@5 & R@10 \\
                \midrule
                MLM+ITM  & 65.2 & 86.4 & 91.5 & 47.2 & 73.1 & 81.3 \\
                MLM+ITM+Pixel & 67.4 & 88.2 & 93.9 & 49.5 & 75.9 & 84.4 \\
                \midrule
                MLM+ITM+Pixel+MIM &  70.1 & 90.2 & 95.3 &  52.4 & 77.8 & 86.0 \\
                \bottomrule
            \end{tabular}
            }
	\end{center}
	\vspace{-3mm}
\end{table}

\subsection{Evaluation on VQA, NLVR and VE}
Following previous approaches~\cite{chen2020uniter,li2021align}, we further report performances on various other vision-language tasks such as VQA, NLVR and SNLI-VE. It's worth noting that some results are not directly comparable as UNITER additionally uses out-of-domain data, OSCAR leverages additional object tags and ~\cite{gan2020large} with adversarial data augmentation. Nevertheless, we observe competitive performance of our method on all tasks across different datasets in Table \ref{tbl:vqa_nlvr_ve}. Specifically, we cast VQA as an answer generation problem~\cite{li2021align}, where we finetune an auto-regressive text decoder which receives inputs from the multi-modal fusion layer. NLVR$^2$ and SNLI-VE are cast as binary classification and three-way classification problem respectively. The input to NLVR$^2$ is a pair of images and a text description, involving comprehensive pairwise relationship reasoning~\cite{suhr2018corpus} such as co-reference, comparisons, negation, coordination etc. We find our approach performing uniformly better than prior approaches on the two tasks. Compared to Image/Text retrieval, VQA, NLVR$^2$ and SNLI-VE require comprehensive reasoning over the relationships between image and text descriptions, which we have explicitly modeled in Equation~\ref{eqn:codebook} as maximizing the reconstruction posterior conditioned on language and visual semantics. We hypothesize that MIM, together with MLM, help strengthen the model's reasoning ability between the two modalities.

\subsection{Ablation Study}
In this section, we do ablation studies on our codebook by manipulating its corresponding loss objectives. Specifically, we conduct zero-shot evaluations on the test set of MSCOCO (5K) for text and image retrieval tasks as they are more reflective of the learned representations. As shown in Table~\ref{table:ablation}, the first row is our ablation study baseline which only reserves the MLM and ITM objectives, as they are commonly used in the literature.  
As shown in Equation~\ref{eqn:codebook}, pixel loss contributes to codebook and visual encoder optimization. A recent approach MAE~\cite{he2021masked} shows that pixel loss is effective for improving unimodal encoder representation. Our experiments in row 2 also confirm the observation. The difference is that ours is a multi-modal framework and we adopt a symmetric encoding/decoding process. By adding masked image modeling on top of codebook optimization, our model accumulates an additional gain of $2.7\%/2.9\%$ in R@1 for TR and IR in the study. 

\begin{figure}[tp]
  \centering
   \includegraphics[width=0.85\linewidth]{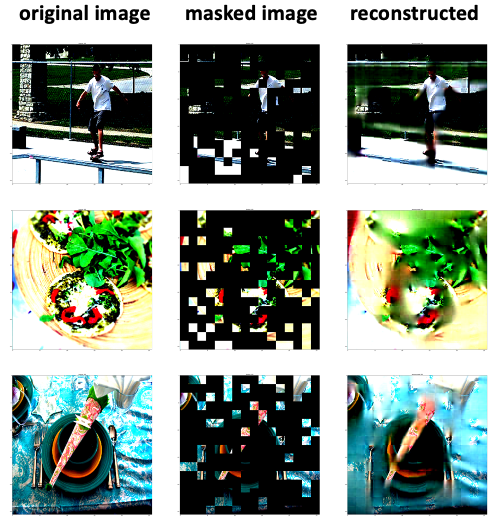}
   \caption{Reconstruction results for the decoded masked image tokens.\vspace{-3mm}}
   \label{fig:recon}
   
\end{figure}
\begin{figure*}[!t]
  \centering
  \includegraphics[width=0.85\linewidth]{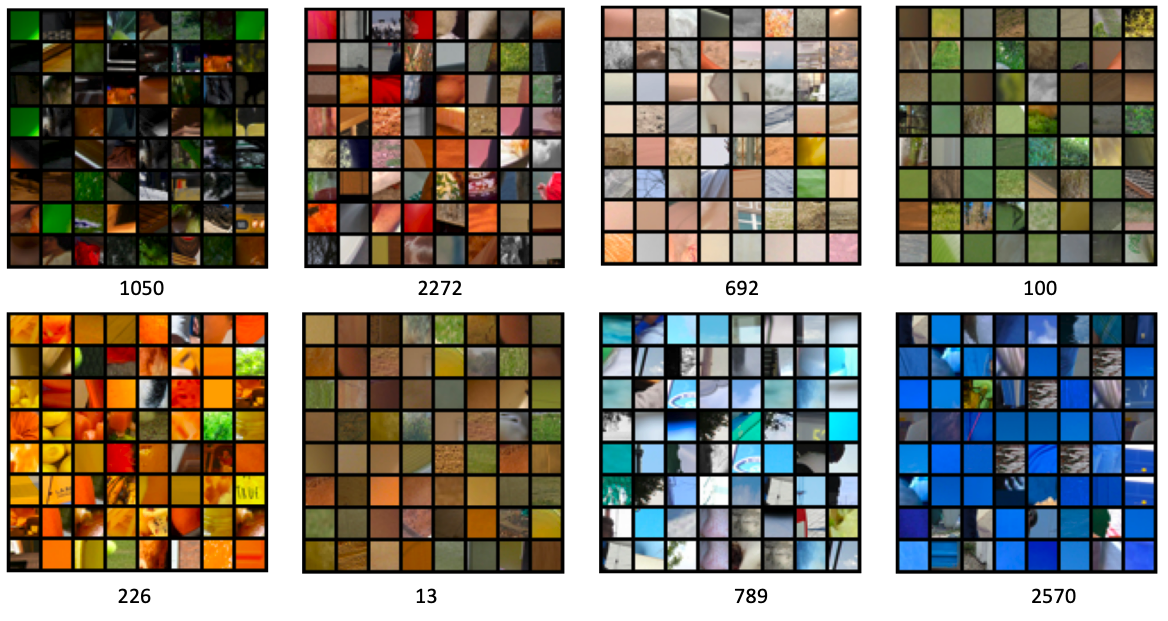}
  \caption{Visualization of image patches that produce the same codeword in different codebooks\vspace{-2mm}.}
  \label{fig:codebook}
     \vspace{-2mm}
\end{figure*}

\begin{figure*}[!ht]
  \centering
   \includegraphics[width=0.85\linewidth]{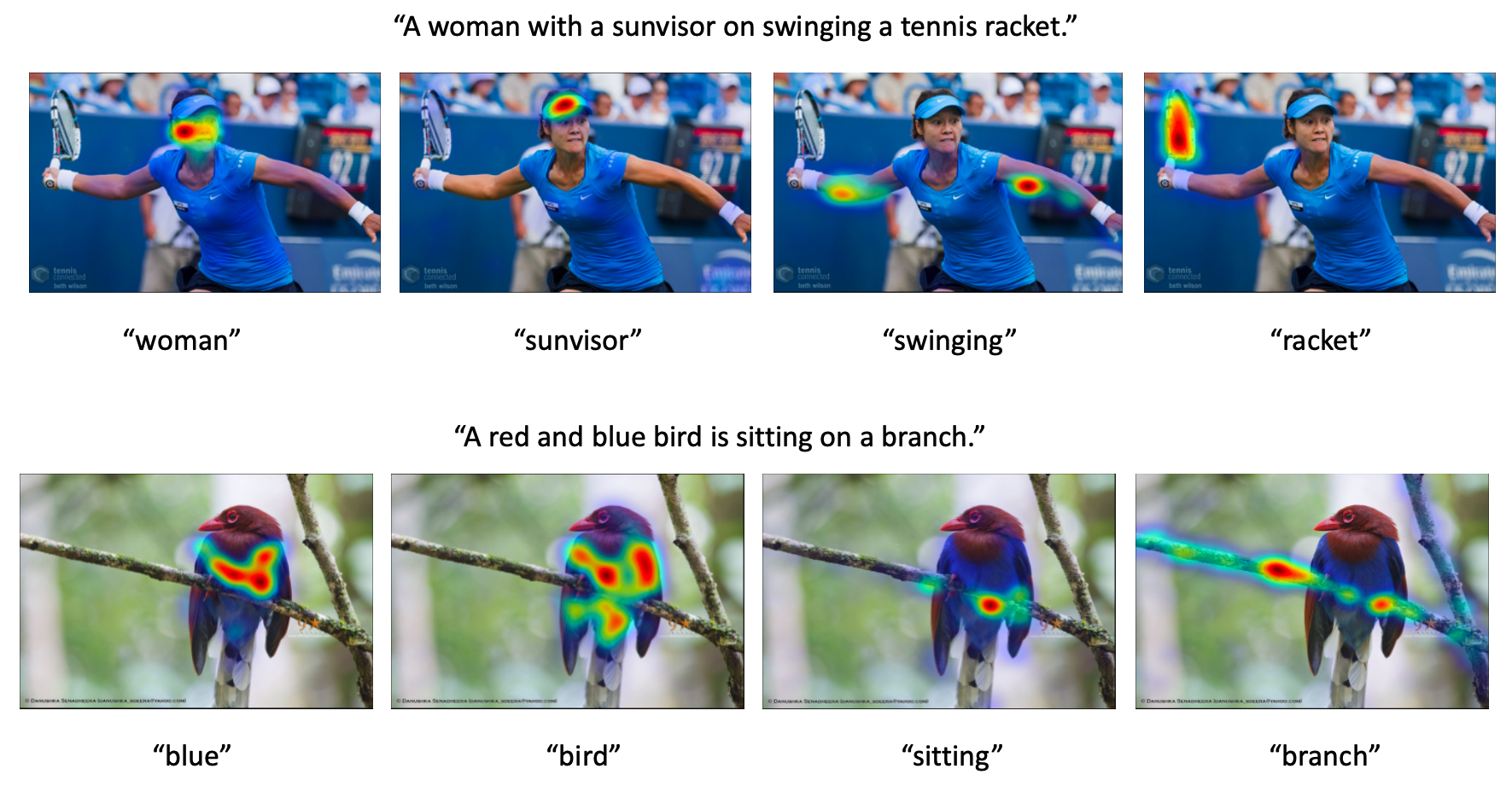}
   \caption{Grad-CAM visualization on the cross-attention maps corresponding to individual words.\vspace{-2mm}}
   \label{fig:attention}
   \vspace{-3mm}
\end{figure*}

\subsection{Codebook Visualization}
\label{sec:codbook_vis}
In this section, we answer the question: \textit{will the learned codebook have semantic meanings?} To answer this, we randomly sample codewords in the codebook, and for each codeword, we group corresponding image patches that are discretized under this codeword index. The visualization of eight random codewords are shown in Figure~\ref{fig:codebook}. We observe that each codeword represents a unique pattern shared among its corresponding patches.

\subsection{Reconstruction Results}
The visualization of the reconstruction results for the masked tokens is displayed in Figure~\ref{fig:recon}. We show the original image (left), the masked image (middle) and our reconstruction (right). The masking ratio is 75\%, leaving only 64 out of 256 patches. The predictions differ plausibly from the original images, showing that the model can generalize. For example in the 3rd row of Figure~\ref{fig:recon}, the reconstructed brush tip is pink, as opposed to green in the original image. Our hypothesis is that the model infers the color from neighboring pink patches whereas the green patches were invisible (masked) as shown in the middle.

\subsection{Cross-attention Visualization}
We visualize the cross-attention maps using Grad-CAM\cite{selvaraju2017grad} to provide qualitative assessment of our approach. Figure~\ref{fig:attention} shows that our method is able to associate language with ``regions of interest'' by attending to meaningful objects and locations, visually reflecting the quality of our model in multimodal alignment. For example, in Figure~\ref{fig:attention}, the model is able to associate noun words such as ``woman", ``sunvisor", ``bird", ``branch" with the correct regions in the image. At the meantime, for verbs such as ``swinging" and ``sitting", the model is able to attend to meaningful semantics by fixating on the arms performing the action and the twig where the bird's leg touches.

\section{Conclusion}
Aligning signals from visual and language modalities is not easy in vision-language representation learning due to the mismatch, where visual signal is continuous and high-dimensional as opposed to language which is discretized in nature and semantics-rich. We propose \textbf{CB-ViLA} to discover the visual semantics via joint-training of a codebook, which in turn helps bridge the semantic gap by discretizing the visual signals. In this way, we symmetrically derive an equivalent of ``MLM'' for the vision side (i.e., MIM), which is conditioned on text and self-supervised. Ablation studies and downtream evaluations reveal that discretizing visual signals with visual semantic codebook facilitates alignment and multi-modal interactions with text signals. We hope to inspire more work in this direction. 
\newpage


\end{document}